\documentclass[final]{cvpr}

\usepackage{times}
\usepackage{epsfig}
\usepackage{graphicx}
\usepackage{amsmath}
\usepackage{amssymb}
\usepackage{siunitx}
\usepackage{bbding}

\usepackage[pagebackref=true,breaklinks=true,colorlinks,bookmarks=false]{hyperref}

\def\cvprPaperID{****} % *** Enter the CVPR Paper ID here
\def\confYear{CVPR 2021}

\begin{document}

%%%%%%%%% TITLE
\title{Deep Layout of Custom-size Furniture through Multiple-domain Learning}

\author{Xinhan Di \Envelope $^{1}$\\
IHome Company Nanjing\\
{\tt\small deepearthgo@gmail.com}
\and
Pengqian Yu $^{2}$\\
IBM Research Singapore\\
{\tt\small peng.qian.yu@ibm.com}
\and
Danfeng Yang $^{3}$\\
Radio Company Nanjing\\
{\tt\small breezeydf@gmail.com}
\and
Hong Zhu $^{4}$\\
IHome Company Nanjing\\
{\tt\small jszh0825@gmail.com}
\and
Changyu Sun $^{5}$\\
Huazhong University of Science and Technology\\
{\tt\small m202072256@hust.edu.cn}
\and
YinDong Liu $^{6}$\\
Tongji University\\
{\tt\small 2030237@tongji.edu.cn}
}

\maketitle

%%%%%%%%% ABSTRACT
\begin{abstract}
In this paper, we propose a multiple-domain model for producing a custom-size furniture layout in the interior scene. This model is aimed to support professional interior designers to produce interior decoration solutions with custom-size furniture more quickly. The proposed model combines a deep layout module, a domain attention module, a dimensional domain transfer module, and a custom-size module in the end-end training. Compared with the prior work on scene synthesis, our proposed model enhances the ability of auto-layout of custom-size furniture in the interior room. We conduct our experiments on a real-world interior layout dataset that contains $710,700$ designs from professional designers. Our numerical results demonstrate that the proposed model yields higher-quality layouts of custom-size furniture in comparison with the state-of-art model. The dataset and codes can be found at \url{https://github.com/CODE-SUBMIT/dataset2}.
\end{abstract}

%%%%%%%%% BODY TEXT
\section{Introduction}

People spend plenty of time indoors such as the bedroom, living room, office, and gym. Function, beauty, cost, and comfort are the keys to the redecoration of indoor scenes. The proprietor prefers demonstration of the layout of indoor scenes in several minutes nowadays, and online virtual interior tools are becoming useful to help people design indoor spaces. These tools are faster, cheaper, and more flexible than real redecoration in real-world scenes. This fast demonstration is often based on the auto layout of indoor furniture and a good graphics engine. Machine learning researchers make use of virtual tools to train data-hungry models for the auto layout \cite{Dai_2018_CVPR,Gordon_2018_CVPR}. The models reduce the time of layout of furniture from hours to minutes and support the fast demonstration. 

Generative models of indoor scenes are valuable for the auto layout of the furniture. This problem of indoor scenes synthesis is studied since the last decade. One family of the approach is object-oriented which the objects in the space are represented explicitly \cite{10.1145/2366145.2366154,10.1145/3303766,Qi_2018_CVPR}. The other family of models is space-oriented which space is treated as a first-class entity and each point in space is occupied through the modeling \cite{10.1145/3197517.3201362}.

\begin{figure*}
\centering
\includegraphics[height=4.0cm]{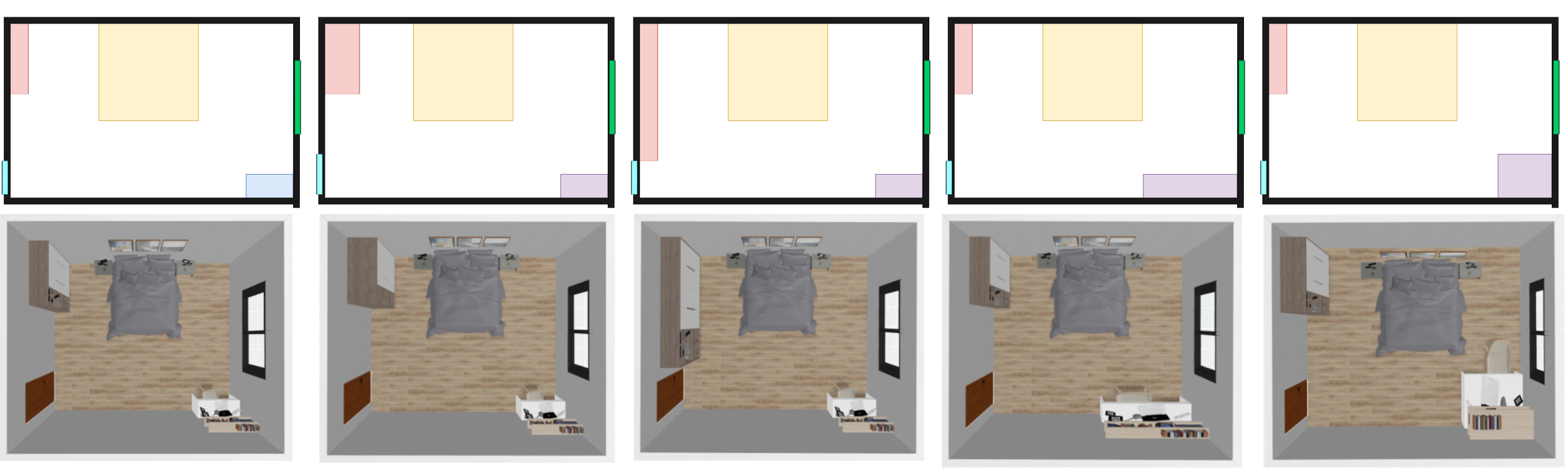}
\caption{Samples of the layout of rooms with custom-size furniture. There are seven types of rooms with custom-size furniture. For example, in the bedroom, the size of the custom cabinet is in five types including the default type. In detail, the width of the custom cabinet grows to the left and the right. The length of the custom cabinet grows to the up and the bottom.}
\label{fig1}
\end{figure*}

Deep generative models are used for efficient generation of indoor scenes for auto-layout recently. These deep models further reduce the time from minutes to seconds. The variety of the generative layout is also increased. The deep generative models directly produce the layout of the furniture given an empty room. However, the direction of a room is diverse in the real world. The south, north, or northwest directions are equally possible. The layout for the real indoor scenes is required to meet with different directions as illustrated in Figure \ref{fig1}. 

Motivated by the above-mentioned challenge, we propose a multiple-domain model for producing custom-size furniture layout in the interior scene. The model yields a design layout of furniture when the size of the furniture is customized. This proposed model aims to support the interior designers to produce decoration solutions with customized furniture in the industrial process. In particular, this proposed adversarial model consists of several modules including a deep layout module, a domain attention module, a dimensional domain transfer module, and a custom-size module in the end-end training.

This paper is organized as follows: the related work is discussed in Section 2. Section 3 introduces the problem formulation. The methods of the proposed adversarial model are in Section 4. The proposed dataset is in Section 5. The experiments and comparisons with the state-of-art models can be found in Section 6. The paper is concluded with discussions in Section 7.

%-------------------------------------------------------------------------
\section{Related Work}

Our work is related to data-hungry methods for synthesizing indoor scenes through the layout of furniture unconditionally or partially conditionally, which we review in the following.

\subsection{Structured data representation}
Representation of scenes as a graph is an elegant methodology since the layout of furniture for indoor scenes is highly structured. In the graph, semantic relationships are encoded as edges, and objects are encoded as nodes. A small dataset of annotated scene hierarchies is learned as a grammar for the prediction of hierarchical indoor scenes \cite{10.1145/3197517.3201362}. Then, the generation of scene graphs from images is applied, including using a scene graph for image retrieval \cite{Johnson_2015_CVPR} and generation of 2D images from an input scene graph \cite{Johnson_2018_CVPR}. However, the use of this family of structure representation is limited to a small dataset. In addition, it is not practical for the auto layout of furniture in the real world. 

\subsection{Indoor scene synthesis}
Early work in the scene modeling implemented kernels and graph walks to retrieve objects from a database \cite{Choi_2013_CVPR,Dasgupta_2016_CVPR}. The graphical models are employed to model the compatibility between furniture and input sketches of scenes \cite{10.1145/2461912.2461968}. However, these early methods are mostly limited by the scenes size. It is therefore hard to produce good-quality layout for large scene size. With the availability of large scene datasets including SUNCG \cite{Song_2017_CVPR}, more sophisticated learning methods are proposed as we review them below.

\subsection{Image CNN networks}
An image-based CNN network is proposed to encoded top-down views of input scenes, and then the encoded scenes are decoded for the prediction of object category and location \cite{10.1145/3197517.3201362}. A variational auto-encoder is applied to the generation of scenes with the representation of a matrix. In the matrix, each column is represented as an object with location and geometry attributes \cite{10.1145/3381866}. A semantically-enriched image-based representation is learned from the top-down views of the indoor scenes, and convolutional object placement priors are trained \cite{10.1145/3197517.3201362}. However, this family of image CNN networks is not applicable to the furniture whose size is customized.

\subsection{Graph generative networks}
A significant number of methods have been proposed to model graphs as networks \cite{DBLP:journals/corr/abs-1709-05584,4700287}. The family for the representation of indoor scenes in the form of tree-structured scene graphs is studied. For example, Grains \cite{10.1145/3303766} consists of a recursive auto-encoder network for the graph generation and it is targeted to produce different relationships including surrounding and supporting. Similarly, a graph neural network is proposed for scene synthesis, where the edges are represented as spatial and semantic relationships of objects \cite{10.1145/3197517.3201362} in a dense graph. Both relationship graphs and instantiation are generated for the design of indoor scenes. The relationship graph helps to find symbolical objects and the high-lever pattern \cite{10.1145/3306346.3322941}. However, the customized size of the furniture is hard to be represented in the abstract graph or relationship graph. 

\subsection{CNN generative networks}
The layout of indoor scenes is also explored as the problem of the generation of the layout. Geometric relations of different types of 2D elements of indoor scenes are modeled through the synthesis of layouts. This synthesis is trained through an adversarial network with self-attention modules \cite{DBLP:journals/corr/abs-1901-06767}. A variational autoencoder is proposed for the generation of stochastic scene layouts with a prior of a label for each scene \cite{Jyothi_2019_ICCV}. However, the generation of the layout is limited to finished furniture of the indoor scenes where the size of the furniture can not be changed.

\section{Problem Formulation}
\begin{figure*}
\centering
\includegraphics[height=7.0cm]{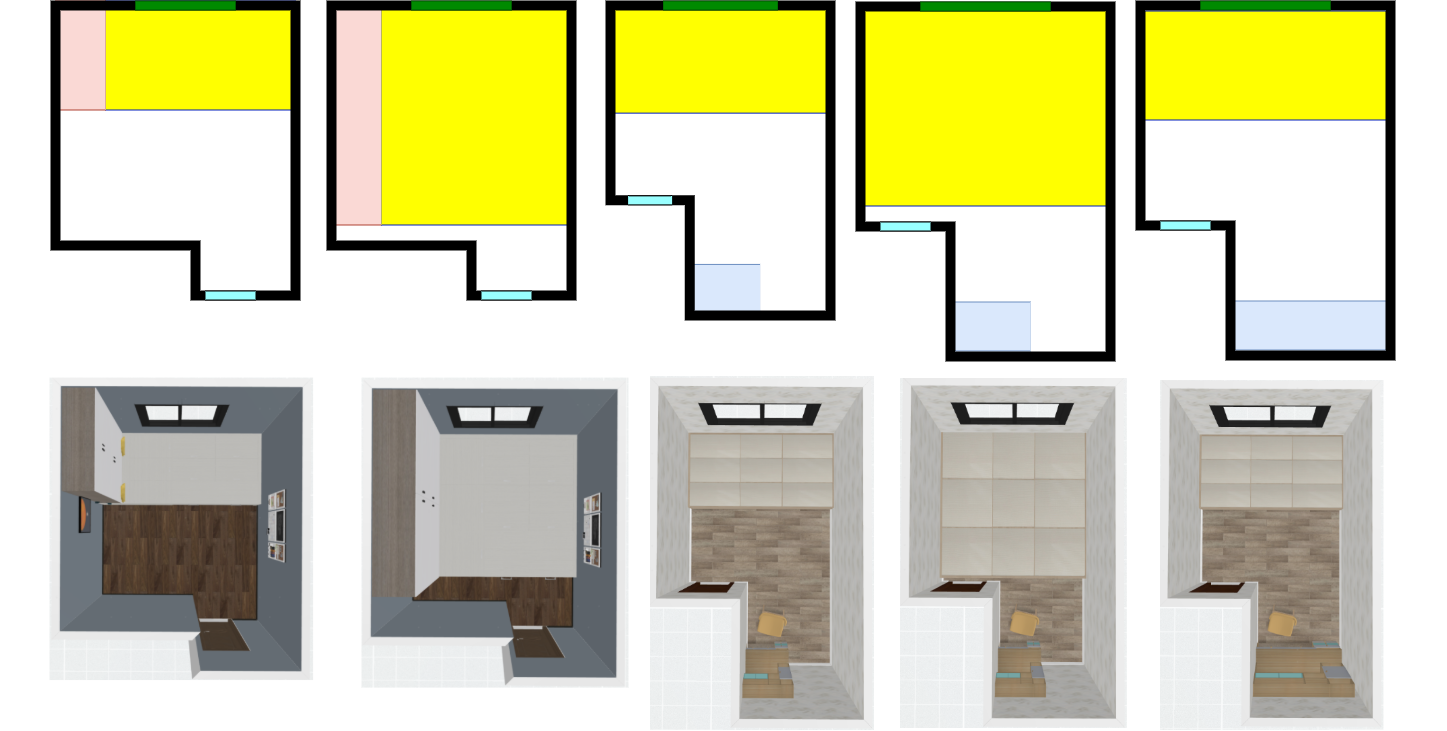}
\caption{Samples of the layout for the Tatami. The size of two custom-size furniture including the working desk and tatami are illustrated. }
\label{fig2}
\end{figure*}

We focus on the problem of layout of custom-size furniture in the indoor scenes. We are given a set of indoor scenes ${(x_{1},y_{1}),\dots,(x_{N},y_{N})}$ where $N$ is the number of the scenes, and $x_{i}$ is an empty indoor scene with basic elements including walls, doors and windows. $y_{i}$ is the corresponding layout of the furniture for $x_{i}$. Each $y_{i}$ contains the elements ${p_{j},s1_{j},s2_{j}}$: $p_{j}$ is the position of the $j$-th furniture; $s1_{j}$ is the default size of the $j$-th furniture; and $s2_{j}$ is the range of size of $j$-th furniture. Each furniture is in the indoor scene $y_{i}$. Figure \ref{fig2} illustrates several instances of different sizes of customized furniture.

Most of the furniture is the customized furniture, where the default size of the $j$-th furniture $s1_{j}$ contains three default value $(\text{length}_{j},\text{width}_{j},\text{height}_{j})$. The range of size $s2_{j}$ contains the ranges of length ${\text{length}_{j}^{\text{min}},\text{length}_{j}^{\text{max}}}$, width ${\text{width}_{j}^{\text{min}},\text{width}_{j}^{\text{max}}}$ and height ${\text{height}_{j}^{\text{min}},\text{height}_{j}^{\text{max}}}$ of the $j$-th furniture.  We define a multiple domain model $M$ such that $y_{\text{pre}} = M(x_{\text{in}},l_{\text{in}})$, and given an empty room $x_{\text{in}}$ with walls, windows and doors, a label $l_{\text{in}}$ representing the dimensional requirements of each customized furniture, the model $M$ produces the layout $y_{\text{pre}}$ including the position, size of each furniture.

\section{Methods}
In this section, we propose a multiple-domain model to produce the layout of each room following the given dimensional requirements of the customized furniture. This model is an end-to-end training model that contains the following modules: a generative layout module, a domain-attention module, a dimensional domain transfer module, and a custom-size module. 

\begin{figure*}
\centering
\includegraphics[height=4.5cm]{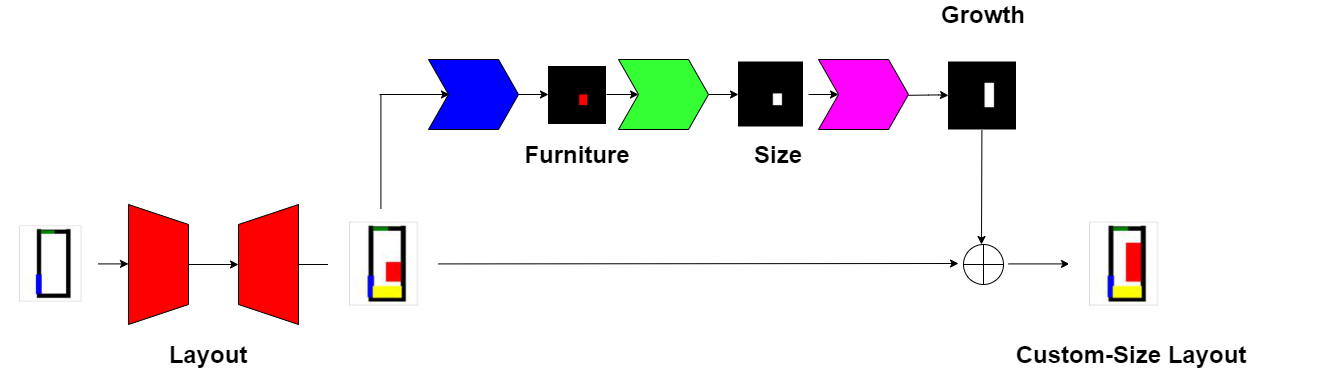}
\caption{Architecture of the proposed model which consists of a generative layout module (in red), a domain-attention module (in blue), a dimensional domain transfer module (in green), and a custom-size module (in purple).}
\label{fig3}
\end{figure*}

In the learning process of this proposed model, we first transfer the domain from the layout to the customized furniture. Secondly, we transfer from the domain of customized furniture to the domain of dimensional requirements of the furniture. Thirdly, we transfer from the dimensional requirements of the furniture back to the domain of the layout. The proposed multiple-domain model as well as the modules are shown in Figure \ref{fig3}. In the following, we will discuss those modules as well as the training objective. 

\subsection{Generative layout module}
\begin{figure}
\centering
\includegraphics[height=2.5cm]{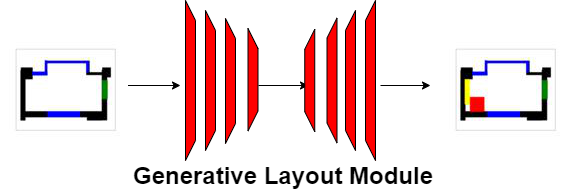}
\caption{Architecture of the proposed generative layout module with the given floor-plan of an empty room as input. It produces the layout of the customised furniture with default size. The categories of the furniture are shown in different colors.}
\label{fig4}
\end{figure}

In this generative layout module, the generation part $g_{1}$ gets the input of a rendered image of an empty room and produces the layout of the customized furniture with default size $y_{i}$. The discriminator part $d_{1}$ determines whether the generated layout is real as illustrated in Figure \ref{fig4}. 

\subsection{Domain attention module}
\begin{figure}
\centering
\includegraphics[height=3.0cm]{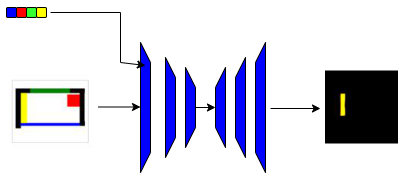}
\caption{Architecture of the domain attention module with the given predicted layout as input. It produces the position, size and category of the customized furniture in the local domain of the customized furniture.}
\label{fig5}
\end{figure}

This attention module $trans_{1}$ transfers the domain from the global layout to each local customized furniture. Given the layout which contains the size and location of walls, windows, doors, and furniture, $trans_{1}$ transfers to the domain of each customized furniture $tp_{1}$ which contains size, location, category of the customized furniture in the layout. This module $trans_{1}$ takes the input of the layout $y_{i}$ and the category code of customized furniture $label_{1}$ for the learning of the domain transfer, as illustrated in Figure \ref{fig5}.

\subsection{Dimensional domain transfer module}
\begin{figure}
\centering
\includegraphics[height=2.0cm]{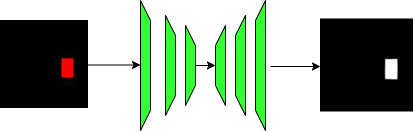}
\caption{Architecture of the dimensional domain transfer module with the predicted customized furniture in the local domain as input. It produces the position and size of the customized furniture in the dimensional domain.}
\label{fig6}
\end{figure}

This dimensional domain module $trans_{2}$ transfers the domain from the local customized furniture to the dimensional domain. This dimensional domain only contains the size of the customized furniture without the location, category of the furniture. This module $trans_{2}$ gets the input of the furniture $tp_{1}$ in the domain of the local customized furniture, then transfers to the dimensional domain $tp_{2}$ as illustrated in Figure \ref{fig6}.   

\subsection{Custom-size module}
\begin{figure}
\centering
\includegraphics[height=4.5cm]{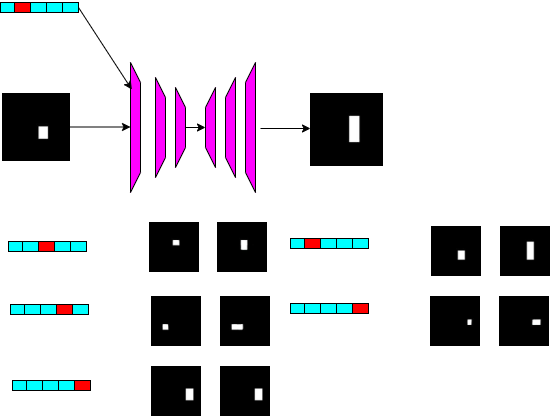}
\caption{Architecture of the custom-size module with the default size of the customized furniture in the dimensional domain. This module produces the growth of size of the customized furniture.}
\label{fig7}
\end{figure}

In the dimensional domain $tp_{2}$, this custom-size module $g_{2}$ produces the size of the customized furniture with given code $label_{2}$ of the required size. Given $label_{2}$ and default size $ls_{1}$ of the customized furniture in $tp_{2}$, the produced size $ls_{2}$ is also in the domain $tp_{2}$ as illustrated in Figure \ref{fig7}. 

%Problem Formulation
%Method
%1.Generative Layout
%2.Domain Attention Module
%3.Domain Transfer
%4.Custom-size Module
%Dataset

\subsection{Training objectives}
We let $g_{1}$ denote the generator of the generative layout module, $d_{1}$ denote its discriminator. We further let $trans_{1}$ denote the domain attention module, $trans_{2}$ denote the dimensional domain transfer module, and $g_{2}$ denote the custom-size module. Given a rendered image of the indoor scene, $x_{i}$ is of size $H \times W \times 3$, where $H$ and $W$ denote the height and width of the rendered image. In addition, $y_{i}$ is also given as the layout of furniture in the scene $x_{i}$: it contains the location and the default size of the customized furniture. Besides of the layout for the customized furniture with default size, a variety of layouts for the customized furniture with different size $y_{i}^{j}, j={1,2,\dots,n_{i}}$ is given. In each layout, $y_{i}^{j}$, the code of the size of the customized furniture $label_{2}$, and the code of the category of the customized furniture $label_{1}$ are also given. 

In the learning of this multiple-domain models, $g_{1}$ is first learned to produce the layout $y_{i}$ given $x_{i}$. Secondly, $trans_{1}$ is learned to produce $tp_{1}$ in the domain of local customized furniture given the layout $y_{i}$ and the code $label_{1}$ of the category of the customized furniture. Thirdly, $trans_{2}$ is learned to produce $tp_{2}$ in the dimensional domain of the furniture given $tp_{1}$. Next, $g_{2}$ is learned to produce the required size $ls_{2}$ in the domain $tp_{2}$ given the code of the size $label_{2}$. Note that $g_{1}$ produces the layout $y_{i}$ where each customized furniture has default size $ls_{1}$. The layout $y_{i}^{j}$, where the customized furniture receive the required size given the code of the size and category $label_{1}$ and $label_{2}$, is produced after $ls_{2}$ is obtained.

\begin{figure*}
\centering
\includegraphics[height=8.0cm]{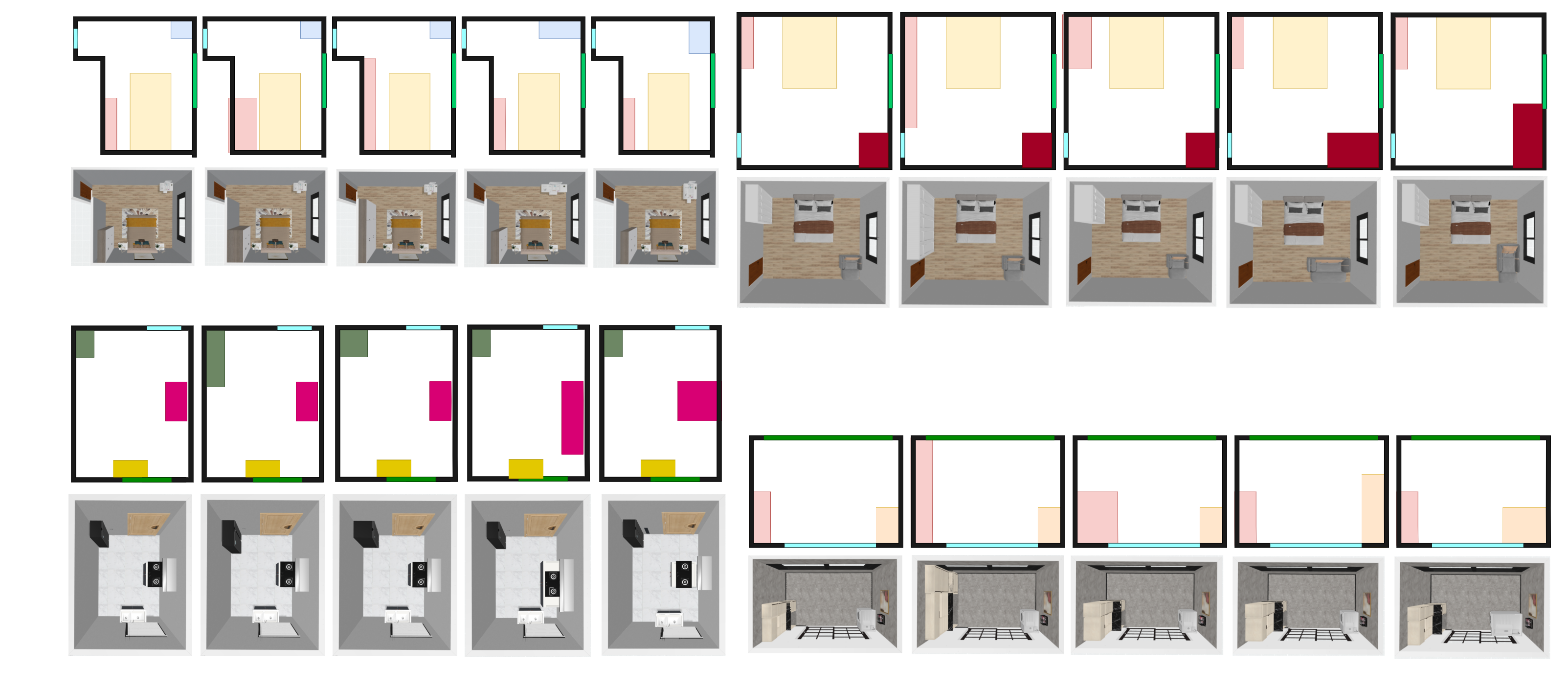}
\caption{The layout samples and the corresponding rendered scenes are illustrated. For each sample, the layout samples including the position, the direction, and the size of each customized furniture with different dimensional sizes are presented.}
\label{fig8}
\end{figure*}

\subsubsection*{Generative layout module training}
To train the discriminator network $d_{1}$, the discriminator loss $L_{D}^{r}$ has the following form:
\begin{equation}
L_{D} = -(1-y_{1n})\log(D(P_{n1}^{0})) + y_{1n}\log(D(P_{n1}^{1}))    
\end{equation}
where $y_{1n}=0$ if sample $P_{n}^{0}$ is drawn from the generator $g_{1}$, and $y_{1n}=1$ if the sample $P_{n}^{1}$ is from the ground truth $y_{i}$. Here, $P_{n}^{0}$ denotes the rendered layout image generated from the generator where all customized furniture has default size $ls_{1}$, and $P_{n}^{1}$ denotes the rendered ground truth layout $y_{i}$ where all customized furniture has default size $ls_{1}$.

To train the generator network $g_{1}$, the generator loss $L_{G}^{r}$ has the following form:
\begin{equation}
L_{G} = L_{g} + \lambda_{adv}L_{adv}
\end{equation}
where $L_{g}$ and $L_{adv}$ denote the generation loss and the adversarial loss, respectively. Here $\lambda_{adv}$ is the constant for balancing the multi-task training. The generation loss measures the difference between $g1(x_{i})$ and $y_{i}$.

\subsubsection*{Domain attention module training} 
To train the domain attention module $trans_{1}$, the first domain transfer loss $L_{trans_1}$ has the following form:
\begin{equation}
L_{trans_1} = M(trans_1(g1(x_{i}),label_{1}^{i})-tp_{1gt}^{i})
\end{equation}
where $M$ is a common norm function, $label1_{i}$ is the code of the categories of the customized furniture in $y_{i}$, and $tp_{1gt}$ is the ground truth in the domain of the local customized furniture corresponding to the $label_{1}^{i}$.

\subsubsection*{Dimensional domain transfer module training} 
To train the dimensional domain transfer module, the second domain transfer loss $L_{trans_2}$ has the following form:
\begin{equation}
L_{trans_2} = M(trans_2(trans_1(g1(x_{i}),label_{1}^{i}))-tp_{2gt}^{i})
\end{equation}
where $M$ is a common norm function, and $tp_{2gt}$ is the ground truth in the dimensional domain of the furniture.

\subsubsection*{Custom-size module training} 
To train the custom-size module, the size loss $L_{size}$ has the following form:
\begin{equation}
L_{size} = M(g_{2}(trans_2(trans_1(g1(x_{i}),label_{2}^{i})), label_{1}^{i})-ls_{2gt}) 
\end{equation}
where $M$ is a common norm function, $ls_{2gt}$ is the ground truth size of the customized furniture in the dimensional domain of the furniture, and $label_{2}^{i}$ is the given code of size of the customized furniture. The above four modules are then trained end-to-end in the learning process.

\section{Proposed dataset}
In this paper, we propose a dataset of indoor furniture layouts together with an end-to-end rendering image of the interior layout. This layout data is from designers at the real selling end.

\subsection{Interior layouts}
There are $60$ professional designers who work with an industry-lever virtual tool to produce a variety of designs. Among these designs, part of them are sold to the proprietors for their interior decorations. We collect these designs at the selling end and provide $710,700$ interior layouts. Each sample of the layout has the following representation including the categories of the furniture in one room, the position $(x,y)$ and size (height, width, length) of each furniture, the position $(x,y)$ and size (height, width, length) of the doors, windows, and walls in the room. Besides, the category of the finished furniture and customized furniture is given. For the customized furniture, the size (height, width, length) is the default size in the layout $y_{i}$. In the layout $y_{i}^{j}, j \in\{1,2,\dots,n\}$, the size of the customized furniture is different. Figure \ref{fig6} illustrates the samples of layouts adopted from the interior design industry. In particular, there are $2\times10^3$ samples of the layout for balcony, $2.2\times10^4$ samples of the layout for bedroom, $1.85\times10^5$ samples of the layout for kitchen, $4\times10^5$ samples of the layout for bathroom, $8\times10^4$ samples of the layout for living-dining room, $1.7\times10^3$ samples of the layout for study room, and $2\times10^4$ samples of the layout for tatami room. 

\begin{table*}
\centering
\begin{tabular}{|p{2cm}|p{1cm}|p{1cm}|p{1cm}|p{1cm}|p{1cm}|p{1cm}|p{1cm}|p{1cm}|}
 \hline
 \multicolumn{1}{|c|}{}&\multicolumn{2}{|c|}{\text{Mode}}&\multicolumn{2}{|c|}{\text{IoU}}&\multicolumn{2}{|c|}{Transfer}&\multicolumn{2}{|c|}{Size}\\
 \hline
 \hfil Model    &\hfil PlanIT  &\hfil Ours   &\hfil PlanIT &\hfil Ours  &\hfil PlanIT &\hfil Ours  &\hfil PlanIT &\hfil Ours\\
 \hline
\hfil Balcony   &$0.689 \pm 0.123$  &$0.889 \pm 1.05$ &$0.649 \pm 1.97 $ &$0.753 \pm 1.65$ & \centering {-} & $0.845 \pm 1.42 $ & $0.214 \pm 0.07$ & $0.832 \pm 2.14$ \\ \hline
\hfil Bedroom   &$0.691 \pm 0.169$  &$0.815 \pm 1.37$ &$0.597 \pm 1.89 $ &$0.728 \pm 1.95$ & \centering {-} & $0.793 \pm 1.82$ & $0.208 \pm 0.05$ & $0.817 \pm 1.95$\\ \hline
\hfil Kitchen   &$0.608 \pm 0.143$  &$0.821 \pm 1.25$ &$0.632 \pm 2.27 $ &$0.719 \pm 1.29$ & \centering {-} & $0.839 \pm 1.97 $ & $0.217 \pm 0.03$ & $0.826 \pm 1.83$ \\ \hline
\hfil Bathroom  &$0.619 \pm 0.197$  &$0.838 \pm 1.78$ &$0.616 \pm 2.17 $ &$0.728 \pm 1.49$ & \centering {-} & $0.817 \pm 1.39 $ & $0.229 \pm 0.04$ & $0.809 \pm 1.74$ \\ \hline
\hfil Living-ding  &$0.612 \pm 0.172$  &$0.841 \pm 1.75$ & $0.587 \pm 2.83 $ &$0.701 \pm 1.73$ & \centering {-} & $0.827 \pm 1.58 $ & $0.213 \pm 0.09$ & $0.873 \pm 1.51$ \\ \hline
\hfil Study   &$0.629 \pm 0.191$  &$0.874 \pm 1.79$ &$0.591 \pm 1.81 $ &$0.719 \pm 1.85$ & \centering {-} & $0.829 \pm 1.74 $ & $0.229 \pm 0.09$ & $0.871 \pm 2.08$ \\ \hline
\hfil Tatami   &$0.547 \pm 0.263$  &$0.783 \pm 2.35$ &$0.584 \pm 2.24 $ &$0.691 \pm 1.94$ & \centering {-} & $0.837 \pm 1.93 $ & $0.207 \pm 0.08$ & $0.806 \pm 2.17$ \\
 \hline
\end{tabular}
\label{table1}
\caption{Evaluation for our proposed model compared with PlanIT.}
\end{table*}

\subsection{Rendered layouts}
Each layout sample corresponds to the rendered layout images. These images are the key demonstration of the interior decoration. These rendered images contain several views and we collect the top-down view as the rendered view as shown in Figure \ref{fig8}. Therefore, the dataset also contains $710,700$ rendered layouts in the top-down view. Each rendered layout corresponds to a design. Here, all rendered data is produced from an industry-lever virtual tool. 

\begin{figure*}
\centering
\includegraphics[height=7.0cm]{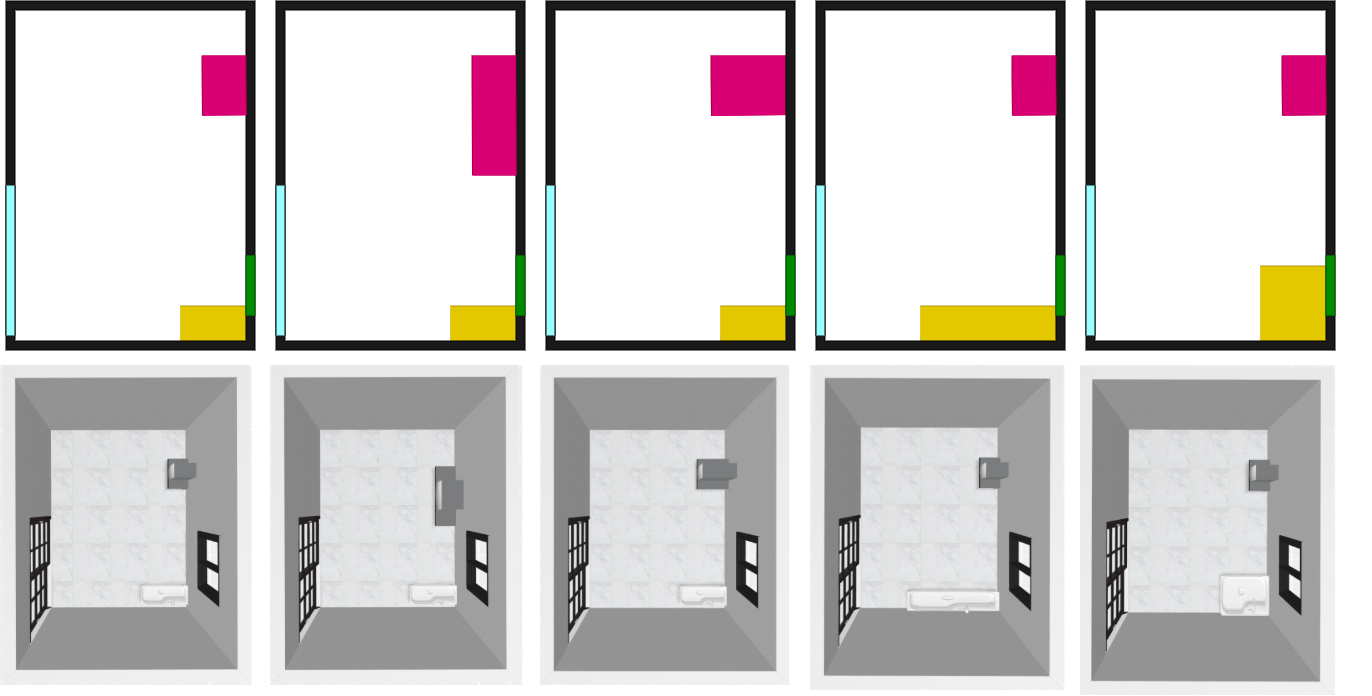}
\caption{Layouts produced by our proposed model for the kitchen.}
\label{fig9_1}
\end{figure*}

\begin{figure*}
\centering
\includegraphics[height=5.0cm]{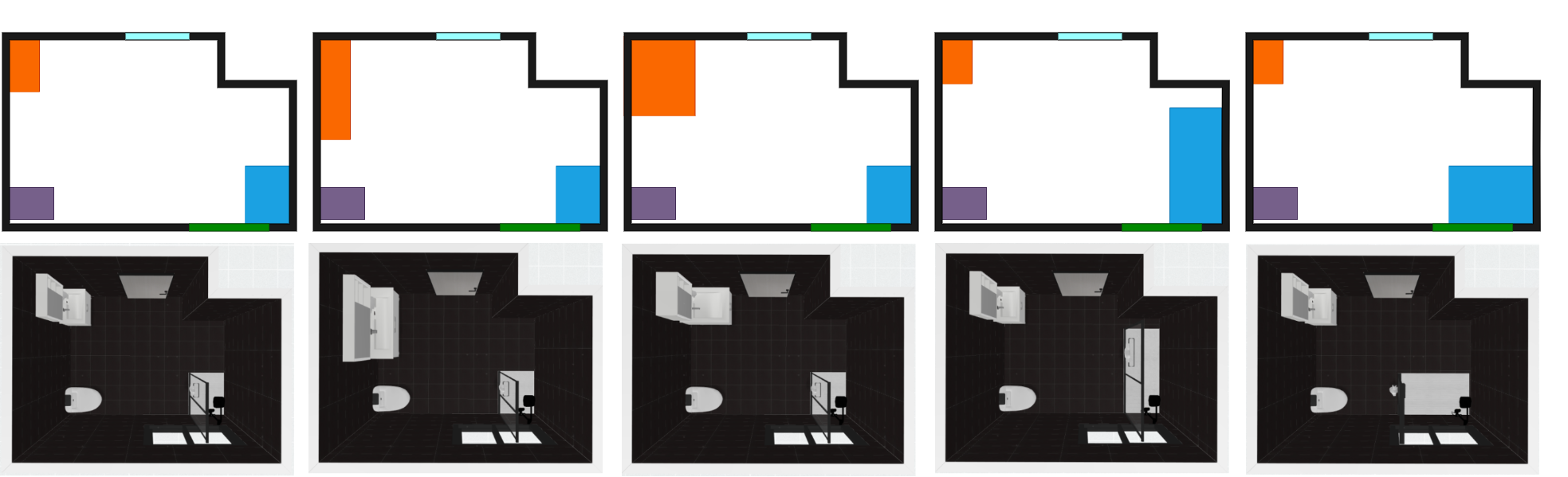}
\caption{Layouts produced by our proposed model for the bathroom.}
\label{fig9_2}
\end{figure*}

\begin{figure*}
\centering
\includegraphics[height=6.3cm]{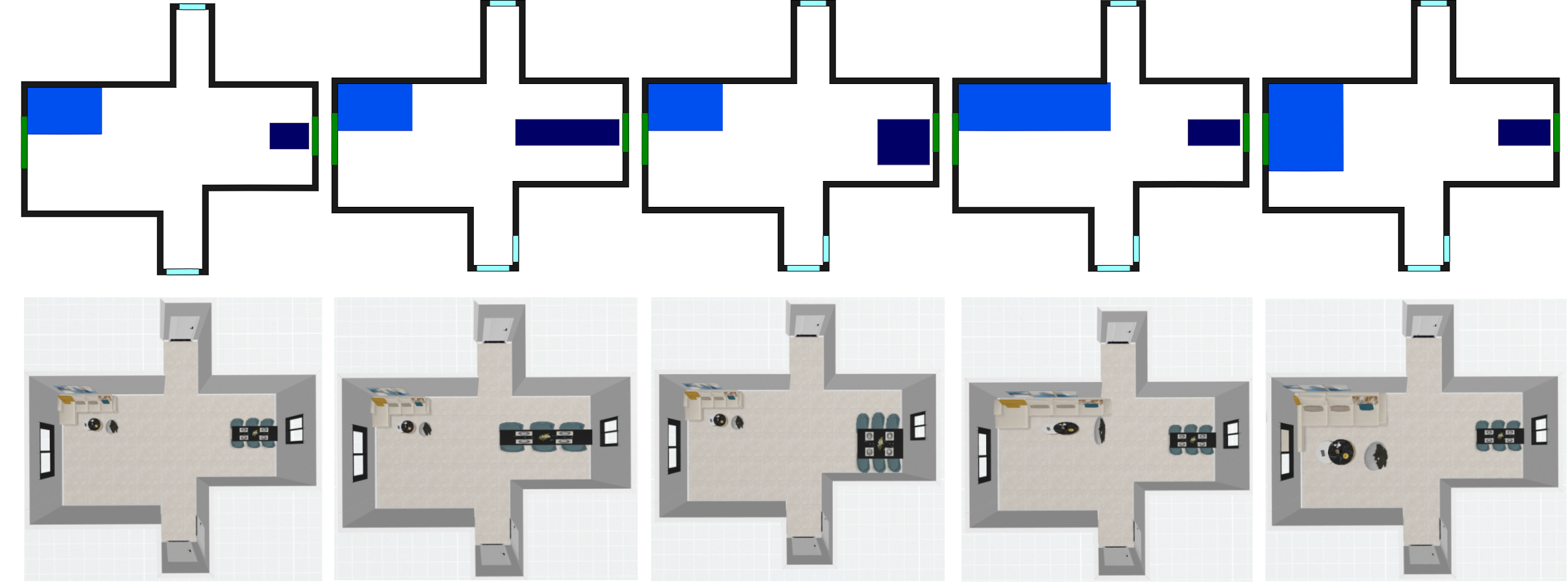}
\caption{Layouts produced by our proposed model for the living-ding room.}
\label{fig9_3}
\end{figure*}

\begin{figure*}
\centering
\includegraphics[height=8.0cm]{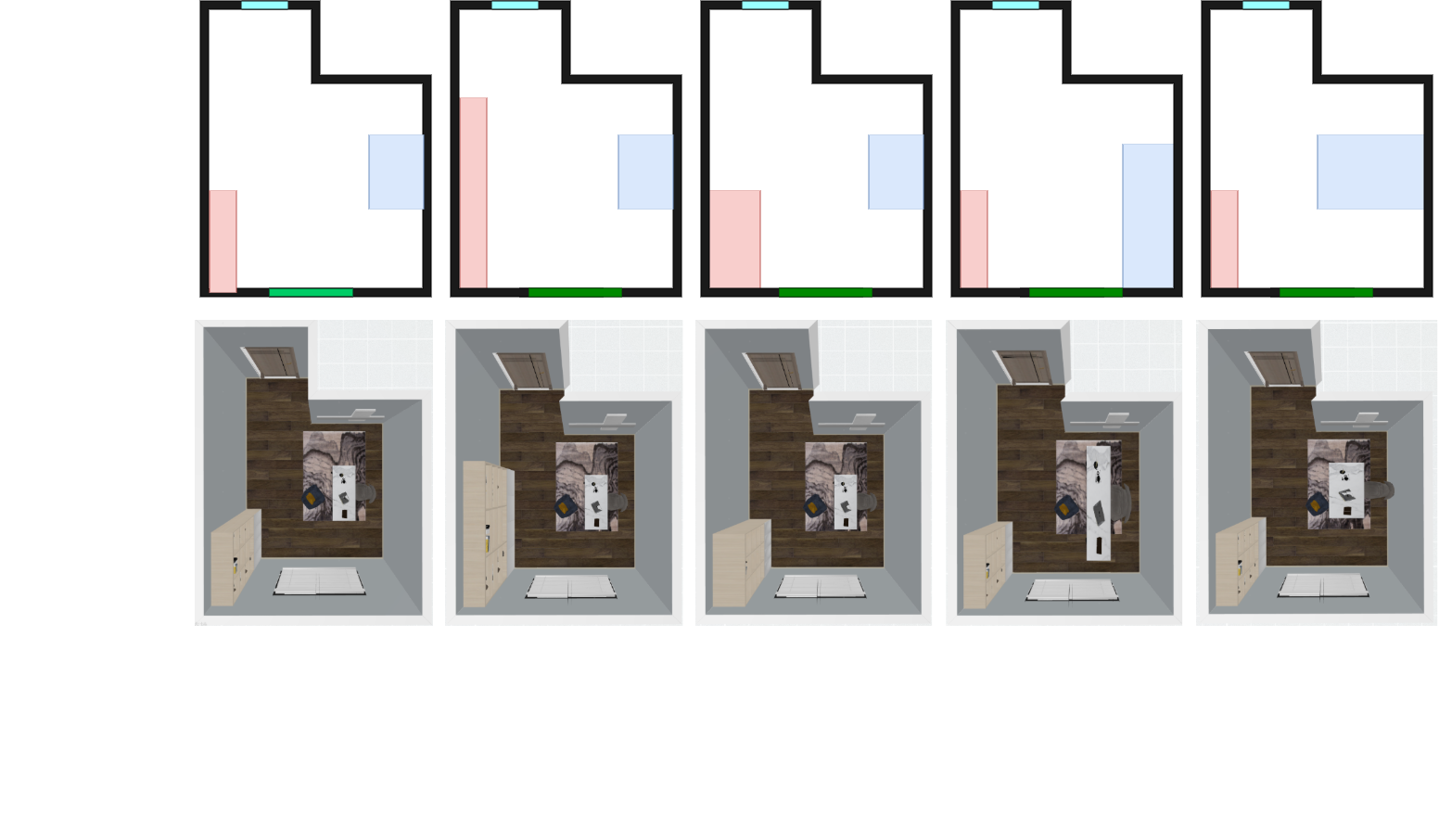}
\caption{Layouts produced by our proposed model for the study room.}
\label{fig9_4}
\end{figure*}

\section{Evaluation}
In this section, we present qualitative and quantitative results demonstrating the utility of our proposed adversarial model for scene synthesis. Six types of indoor rooms are evaluated including the bedroom, the bathroom, the study room, the tatami room, the living-ding room, the kitchen, and the balcony. $90\%$ samples of each room are randomly chosen for training, and $10\%$ samples are used for the test. 

\subsection{Evaluation metrics}
For the task of interior scene synthesis, we apply four metrics for the evaluation. Firstly, we use average mode accuracy for the evaluation. It is to measure the accuracy of the category of furniture for a layout corresponding to the ground truth. This average mode accuracy is defined as
\begin{equation}
\text{Mode} := \frac{\sum_{i=1}^{n}N_{i}^{1}}{\sum_{i=1}^{n}N_{i}^{\text{total}}} 
\end{equation}
where $N_{i}^{\text{total}}$ is the total number of $i$-th category of furniture in the ground truth dataset, and $N_{i}^{1}$ is the number of the $i$-th category of furniture in the generated layout in corresponding with the ground truth. For example, if the $i$-th furniture is in the predicted layout where the ground truth layout also contains this furniture, then it will be calculated. Note that $n$ is the total number of the category of furniture.

Secondly, in order to evaluate the position accuracy of the furniture layout, we apply the classical mean Intersection over Union (IoU) between the predicted box of $i$-th furniture and the ground truth box.

Thirdly, we use the average accuracy in the domain of customized furniture for the evaluation. It is to measure the accuracy of the category of customized furniture in the domain transferring process. This average transferring accuracy is defined as:
\begin{equation}
\text{Transfer} := \frac{\sum_{i=1}^{n}N_{i}^{1}}{\sum_{i=1}^{n}N_{i}^{\text{total}}} 
\end{equation}
where $N_{i}^{\text{total}}$ is the total number of $i$-th category of furniture in the ground truth dataset, and $N_{i}^{1}$ is the number of the $i$-th category of furniture in corresponding with the ground truth in the domain of customized furniture. For example, if the category of the furniture is the same as the ground truth in the domain, it will be calculated. Note that $n$ is the total number of the category of furniture.

Finally, we use the average size accuracy in the dimensional domain for the evaluation. It is to measure the accuracy of the size of customized furniture in the dimensional domain. 
This average transferring accuracy is defined as:
\begin{equation}
\text{Size} := \frac{\sum_{i=1}^{n}N_{i}^{1}}{\sum_{i=1}^{n}N_{i}^{\text{total}}} 
\end{equation}

In the dataset, the size of each customized furniture changes in five modes. Firstly, the size is the same as the default size. Secondly, the width is twice the default width, and the growth direction of the width is the left. Thirdly, the width is twice the default width, and the growth direction of the width is the right. Fourthly, the height is twice the default height, and the growth direction of the width is the up. Finally, the height is twice the default height, and the growth direction of the width is the bottom. 

We compare two baseline models for scene synthesis for seven types of rooms. The results are shown in Figure \ref{fig9_1} -- \ref{fig9_4}. Our model outperforms the state-of-art model in the following aspects. Firstly, the proposed model gives the good size of the customized furniture for the layout while the state-of-art model is unable to give the right size. Secondly, our model predicts a good position of the customized furniture in the layout, while the state-of-art model sometimes predicts an unsatisfied position. 

We also compare with the state-of-art model quantitatively. Four performance metrics for seven types of room are given in Table 1. It can be seen that our model outperforms the state-of-art model in terms of the mode accuracy and the position accuracy.  

\section{Discussion}
In this paper, we present a multiple-learning model to predict the layout interior scene where the furniture is customized. In addition, we propose an interior layouts dataset that all the designs are drawn from the professional designers. The proposed model achieves better performance in comparison with the state-of-art models on the interior layouts dataset. There are several avenues for future work. Our method is currently limited to the generation of layouts for the common rooms, and the layout of other rooms is harder to predict. For example, it is difficult to predict the layout for the luxury bedroom where the bathroom and the cloakroom are also built in the luxury bedroom. Besides, the customized size of the furniture is limited to five types. It is worthwhile to extend our work and study a more general setting where there are more types of customized sizes.

{
\bibliographystyle{ieee_fullname}
\bibliography{egbib}
}

\end{document}